\documentclass[review]{elsarticle}

\usepackage{lineno,hyperref}

\usepackage[T2A]{fontenc}
\usepackage[russian,english]{babel}
\usepackage[usenames]{color}
\usepackage{colortbl}
\usepackage{caption}
\usepackage{subcaption}
\usepackage{multirow}
\usepackage{soul}

\modulolinenumbers[5]

\journal{Neurocomputing}

\begin{document}

\begin{frontmatter}

\title{OFMPNet: Deep End-to-End Model for Occupancy and Flow Prediction in Urban Environment}

\author[affiliation1]{Youshaa Murhij}
\cortext[mycorrespondingauthor]{Corresponding author}
\ead{yosha.morheg@phystech.edu}
\author[affiliation1,affiliation2]{Dmitry Yudin}
\ead{yudin@airi.net}

\address[affiliation1]{Moscow Institute of Physics and Technology, 9 Institutsky per., Dolgoprudny, Russia}
\address[affiliation2]{Artificial Intelligence Research Institute (AIRI), 32 Kutuzovsky Ave., Moscow, Russia}

\begin{abstract}

The task of motion prediction is pivotal for autonomous driving systems, providing crucial data to 
choose a vehicle behavior strategy within its surroundings. Existing motion prediction techniques primarily focus on predicting the future trajectory of each agent in the scene individually, utilizing its past trajectory data.

In this paper, we introduce an end-to-end neural network methodology designed to predict the future behaviors of all dynamic objects in the environment. This approach leverages the occupancy map and the scene’s motion flow.

We are investigating various alternatives for constructing a deep encoder-decoder model called OFMPNet. This model uses a sequence of bird’s-eye-view road images,
occupancy grid, and prior motion flow as input data. The encoder of the model can incorporate transformer, attention-based, or convolutional units. 
The decoder considers the use of both convolutional modules and recurrent blocks.

Additionally, we propose a novel time-weighted motion flow loss, whose application has shown a substantial decrease in end-point error. Our approach has achieved state-of-the-art results on the Waymo Occupancy and Flow Prediction benchmark, with a Soft IoU of 52.1\% and an AUC of 76.75\% on Flow-Grounded Occupancy.

\end{abstract}


\end{frontmatter}


\section{Introduction}

Currently, the autonomous driving task is divided into multiple subtasks, including: perception, motion prediction and planning. 
The motion prediction subtask allows estimating the future positions of the surrounding moving objects by the perception module.
This can ensure the safe movement of the ego-vehicle in such a dynamically changing environment.
Motion prediction has been studied extensively due to the growing interest in autonomous driving. 
It typically takes road map and agent history states as input. 
To encode such scene context, early works typically rasterize them into an image so as to be processed with convolutional neural networks (CNNs) \cite{paravarzar2020motion}.

Recently, the large-scale Waymo Open Motion Dataset (WOMD) \cite{ettinger2021large} is proposed for long-term motion prediction.
To address this challenge, DenseTNT \cite{gu2021densetnt} adopts a goal-based strategy to classify endpoint of trajectory from dense goal points. 
Other works directly predict the future trajectories based on the encoded agent features or latent anchor embedding \cite{paravarzar2020motion}. 
Transformer has been widely applied in natural language processing and computer vision \cite{konev2022motioncnn}. 
DETR \cite{carion2020end} and its follow-up works, especially DAB-DETR \cite{liu2022dab}, consider the object query as the positional embedding of a spatial anchor box\cite{9733973}.MTR adopts a novel Transformer encoder-decoder structures with iterative motion refinement for predicting multimodal future motion \cite{shi2022motion}.

The main modern approach to solving the problem of predicting the movement of vehicles on the road scene is deep neural network methods. 
They provide the best quality metrics at various leading competitions held using NuScenes \cite{nuscenes2019}, Argoverse \cite{Argoverse2}, Waymo \cite{ettinger2021large} open datasets.

\begin{figure}[t]
     \centering
     \begin{subfigure}[b]{0.3\textwidth}
         \centering
         \includegraphics[width=\textwidth]{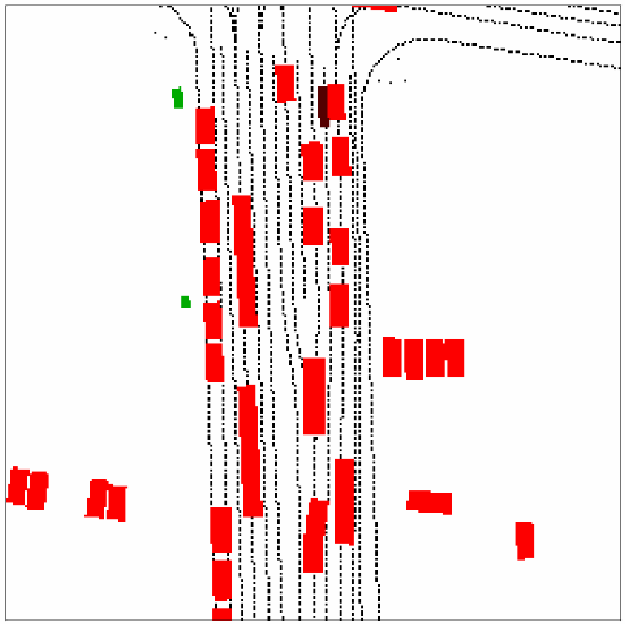}
         \caption{Observed grid}
         \label{fig:gt_obs}
     \end{subfigure}
     \hfill
     \begin{subfigure}[b]{0.3\textwidth}
         \centering
         \includegraphics[width=\textwidth]{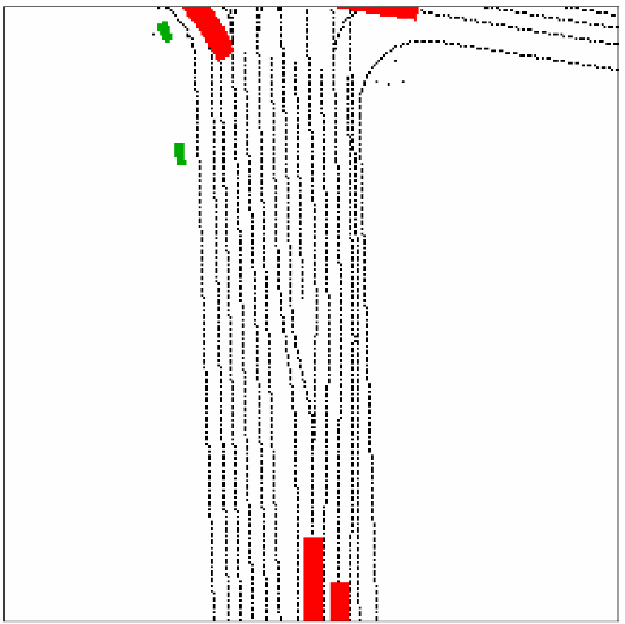}
         \caption{Occluded grid}
         \label{fig:gr_occ}
     \end{subfigure}
     \hfill
     \begin{subfigure}[b]{0.3\textwidth}
         \centering
         \includegraphics[width=\textwidth]{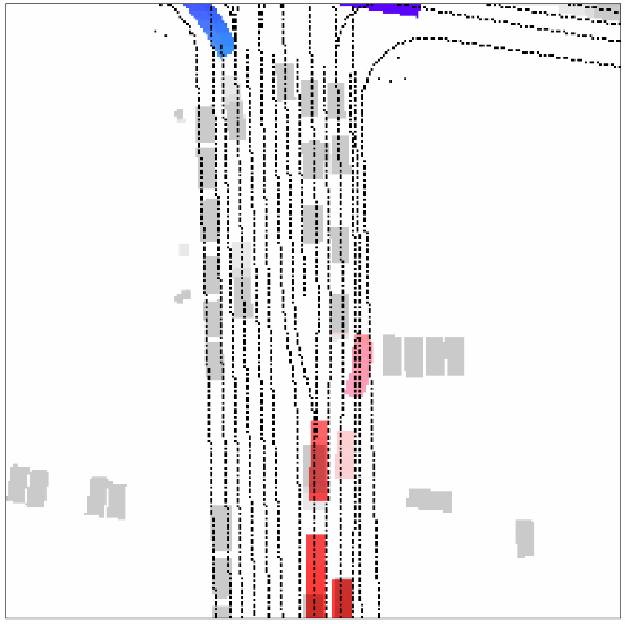}
         \caption{Flow grid}
         \label{fig:gt_flow}
     \end{subfigure}
    \caption{Visualisation of output data for the discussed problem based on Waymo Occupancy and Flow Prediction Challenge data \cite{ettinger2021large}
    }
    \label{fig:gt_vis}
\end{figure}

In this work, we investigate a neural network-based simultaneous solution for three connected tasks for the motion prediction problem presented in \cite{ettinger2021large} (see Figure \ref{fig:gt_vis}):
\begin{itemize}
    \item future occupancy prediction of currently-observed vehicles, 
    \item future occupancy prediction of all vehicles that are not present at the current time step,
    \item future motion flow of all vehicles.
\end{itemize}

We decided to explore the potential of common neural architectures and their modifications in this specific task. Specifically, the ones that include UNet-like backbone, LSTM and SWIN transformer. Additionally, we introduce our improvements and modifications to these models to properly match the task in hand.
Our approach introduces the integration of the SWIN transformer for feature selection with LSTM and sparse operations for feature refinement benefiting from our proposed time-based weight for flow loss.

The first task is the future occupancy prediction of currently-observed vehicles. 
See Figure \ref{fig:gt_vis} (a).
Given histories of all agents over $T_h$ input time steps {$t - T_h, t - (T_h - 1), ..., t$}, 
we need to predict the future occupancy grid as bird's-eye-view (BEV) of all vehicles that are present at the current time step $t$, for $N$ seconds into the future. 
More specifically, the predictions are $N$ occupancy grids, capturing future occupancy of all currently-visible vehicles at $N$ different waypoints. 
Each occupancy grid is a $m x m x 1$ array containing values in the range $[0, 1]$ indicating the probability that some part of some currently-observed vehicle will occupy that grid cell.

In the second task, see Figure \ref{fig:gt_vis} (b), we are given histories of all agents over $T_h$ input time steps {$t - T_h, t - (T_h - 1), ..., t$},
and we need to predict future occupancy of all vehicles that are not present at the current time step $t$, for $N$ seconds into the future. 
The predictions are $N$ occupancy grids, capturing future occupancy of all currently-occluded vehicles at $N$ different waypoints. 
Each occupancy grid is a $m x m x 1$ array containing values in the range $[0, 1]$ indicating the probability that some part of some currently-occluded vehicle will occupy that grid cell.

In the third task, see Figure \ref{fig:gt_vis} (c), we are given histories of all agents over $T_h$  input time steps {$t - T_h, t - (T_h - 1), ..., t$}, 
and we need to predict future motion flow of all vehicles (currently observed or occluded), for $N$ seconds into the future. 
The predictions are $N$ flow fields, capturing future flow of all vehicles at $N$ different waypoints. 
Each flow field is a $m x m x 2$ array containing $(dx, dy)$ values indicating the displacement over 1 second of the vehicle part that occupies that grid cell.

In our article, we focused on building end-to-end neural network models to solve these three tasks and experimenting with their training on the most relevant and up-to-date Waymo Open Motion Dataset \cite{ettinger2021large}.

Our main contributions are listed below:
\begin{itemize}
    \item We present a new deep encoder-decoder model called OFMPNet for occupancy and flow prediction problem. As one of the model options, we proposed a novel architecture with an attention-based, transformer and LSTM units to extract features from historical occupancy/flow grids.
    \item We introduce a time weighted loss as a part of a combination of occupancy-flow losses for multi-task learning. This has demonstrated its effectiveness for the motion flow prediction task.
    \item We train, validate and test our approach on Waymo Open Motion dataset, achieving competing performance in comparison with current state-of-the-art methods.
\end{itemize}

Code of the proposed approach is publicly available at: \\
\url{https://github.com/YoushaaMurhij/OFMPNet}

\section{Related Work}

Transformers gained noticeable success in motion prediction and trajectory estimation tasks as they are computationally efficient and their multi-head attention mechanism is quite effective in time-series and graph interaction encoding \cite{gao2020vectornet}, \cite{9812060}. 
Moreover, Vision Transformer (ViT) \cite{dosovitskiy2020image}, encodes scene features better than CNN-based methods as Vision Transformer provide larger visual reception fields.

Scene Transformers \cite{ngiam2021scene}, \cite{biktairov2020prank}, \cite{djuric2020uncertainty} and \cite{hoermann2018dynamic} predict the behavior of all dynamic objects jointly, producing consistent futures that account for interactions between agents. following language modeling approaches, Scene Transformers use a masking strategy as a query, enabling to predict agent behavior or future trajectory of the autonomous vehicle or the behavior of other agents. 
Scene Transformer architecture employs attention to combine features across road objects, their interactions, and time steps.
 
Another approaches including TrajNet\cite{liu2022strajnet} and HOPE \cite{hu2022hope} propose a multi-modal hierarchical Transformer that combines the vectorized (agent motion) and visual (scene flow, map, and occupancy) information and predicts the flow and occupancy grid of the scene. 
In STrajNet approach, visual and vectorized features are jointly encoded through a multi-stage Transformer and they are fed to a late-fusion module with temporal pixel-wise attention. 
After that, a flow-guided multi-head self-attention block is added to gather the information on occupancy and flow including the mathematical relations between them.
Hierarchical spatial-temporal models in STrajNet benefit from multiple spatial-temporal encoders, which are multi-scale aggregators supported by latent variables, and a sparse 3D decoder. 
HOPE \cite{hu2022hope} uses a combination of losses including focal loss and trace loss to efficiently boost the model training procedure.

Fewer approaches aim to forecast agents' future trajectory based on the occupancy grid representation from the surrounding environment. 
ChauffeurNet \cite{bansal2018chauffeurnet} implements multi-task network for motion prediction based on occupancy input from driving scenarios. 
While DRF-Net \cite{jain2020discrete} relays on auto-regressive sequential prediction to predict a sequence of occupancy residuals.
MP3 \cite{casas2021mp3} forecasts intermediate representations in the form of an online map and the current and future state of dynamic agent as they predict a set of forward motion vectors and associated probabilities per grid cell.
Rules of the Road \cite{hong2019rules} presents a dynamic method to decode likely trajectories from occupancy under a simple motion model with different ways of modelling the future as a distribution over future states using standard supervised learning.

Some approaches aim to use voxelization \cite{murhij2023rethinking} and feature map flow \cite{9892748} to improve generated occupancy heat maps. 
Others \cite{9652040} try to generate an occupancy map of the surrounding space from noisy point clouds obtained from stereo cameras while investigating supervised approaches to train deep networks on unbalanced samples from road scenes.

In \cite{SHARMA2022120}, a survey was conducted on the variety of techniques applied to anticipate pedestrian intention and classified them from multiple perspectives. 
The paper also outlines some newly introduced datasets with added complexities of human behavior on the road. Additionally, it provides a comparative analysis of the performance of pedestrian intention prediction approaches on several benchmark datasets based on various assessment parameters.

In \cite{LIU2022761}, they analyze the impact of traffic accident information on traffic flow and proposes a grey convolutional neural network called G-CNN for predicting traffic flow.

Other works \cite{weng2021ptp} focus on joint object tracking and motion prediction using a feature interaction technique by introducing Graph Neural Networks (GNNs) to capture the way in which agents interact with one another. 
The GNN is able to improve discriminative feature learning for MOT association and provide socially-aware contexts for trajectory prediction. 

\section{Methodology}

\subsection{Problem Description}

The main task is to predict multiple outputs $\mathbf{\hat{Y}}$ that contain the future frames including the future occupancy $\hat{O}^b_k$ of currently observed agents, the future occupancy $\hat{O}^c_k$ of occluded ones (might appear in the future frames) and the future flow $\hat{F}_k$ along both $x$ and $y$ axis at each step $k$ in the future, which represents the motion shifting of grids by the objects, where, $k \in [1, T_f]$ based on the information gathered from the past and current frames of the traffic and scene objects in a certain area. 
The occupancy grids are binary images $\hat{O}^b_k, \hat{O}^c_k \in R^{H \times W \times 1}$. 
The flow is two-channel grid: $ \hat{F}_k = (x,y)_{k-1} - (x,y)_{k} \in R^{H \times W \times 2} $ which could be between $\pm H/2, \pm W/2$. Therefore, the future occupancy can be described as:
\begin{equation}
    O_k = f_w(O_{k-1}, F_{k-1}) \odot O_k.
\end{equation}
where, $f_w$ - is the wrapping function.

\subsection{Input Features}
We refer to the input features as $\mathbf{X}$ for the road and traffic agents. 
We start by constructing the current and previous occupancy grids $O_t, t \in [-T_h, 0]$ and the road map details $M$ including the road lines and traffic lights. 
Additionally, we construct the historical flow $F_h$ from the occupancy grids in the range $-T_h, 0$. 

Next, we encode the trajectory information for the agents in the grid $S = {S_1, S_2, ..., S_n}$, which refers to the motion sequence of states of each agent $s^i_t \in S_i$, where $s^i_t = (x,y,v_x,v_y,\theta,class)$.

Now, we can formulate the task as:
\begin{equation}
    \begin{array}{l}
    \mathbf{\hat{Y}} = f(\mathbf{X}|\theta), \\
    \mathbf{X} = [{O_t|t \in [-T_h,0]};M;S;F_h], \\
    \mathbf{\hat{Y}} = {(\hat{O}^b_k,\hat{O}^c_k,\hat{F}_h)|k \in [1,T_f]}.
    \end{array}
\end{equation}

\begin{figure*}[t]
\begin{center}
\includegraphics[width=1\linewidth]{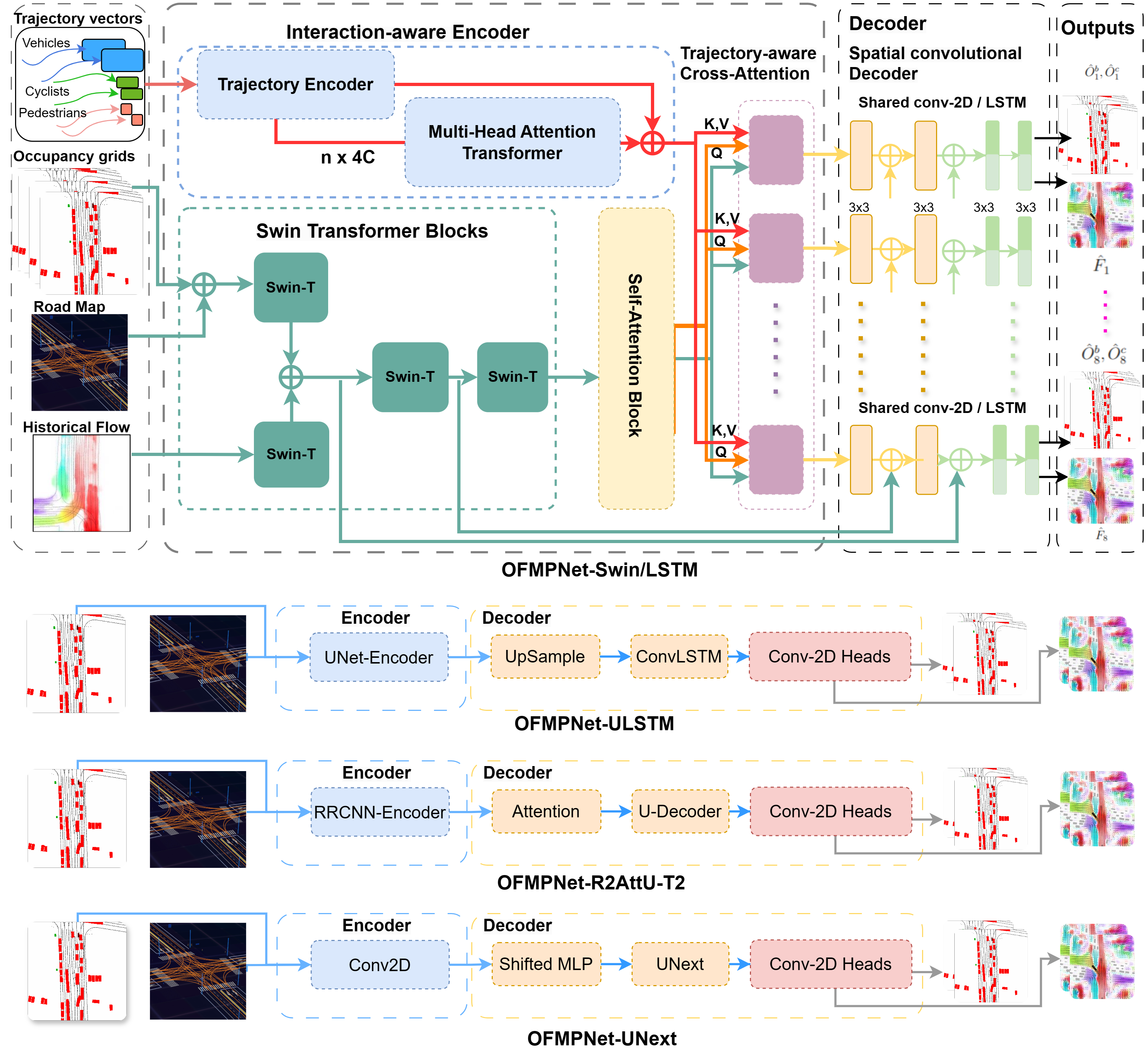}
\end{center}
   \caption{Overview of proposed OFMPNet architectures}
\label{fig:ofmpnet}
\end{figure*}

\subsection{Model Architectures}

To solve the described problem, we have developed three architecture variants of deep encoder-decoder model called OFMPNet, shown in Figure \ref{fig:ofmpnet}:
\begin{itemize}
    \item OFMPNet-Swin,
    \item OFMPNet-ULSTM, 
    \item OFMPNet-R2AttU-T2. 
\end{itemize}

\subsubsection{OFMPNet-Swin} 
Proposed OFMPNet-Swin pipeline consists of two separate encoders, one to encoder agents trajectories $S$ and the other to encode occupancy and flow input $O_t, M, F_h$, which are combined in an early stage. 
We adopt a simple convolutional encoder with Multi-Head Attention \cite{vaswani2017attention} to capture the interaction between the agents for trajectory encoding while in the other encoder we use Multiple Swin-Transformer blocks \cite{liu2021swin} followed by attention mechanism to combine high-level flow and occupancy features.
Next, we feed the combined features to a Cross-Attention block where we also feed the previously encoded trajectory features as keys and values. Finally, we add a residual spacial decoder with temporal shared layers to jointly predict the occupancy and flow grids $\hat{\mathbf{Y}}$.

Let's consider the work of the main modules in more detail.

\textbf{OFMPNet-Swin Encoders.} We use Swin-Transformer block \cite{liu2021swin} to encode the combined occupancy grid maps $O_t$ and the road maps $M$ and another Swin-Transformer block to encode the flow $F_h$, separately.
$O_t$ and $M$ are scaled down by a factor of $4$ using convolution layers with a kernel-size of $4\times4$ and stride of $4$.
We keep the Swin-Transformer settings as described in \cite{liu2021swin} including multi-head window and shifted-window self-attention with encoding bias $B$:
\begin{equation}
 \begin{array}{r}
    MSA(Q, K, V) = (head_1 || ... || head_i) W^O, \\
    head_i = softmax(QK^T / \sqrt{d} + B) V,
 \end{array}
\end{equation}
where, $MSA$: multi-head self-attention, $i$ number of heads; i = [3, 6, 12] and $d$ is the length of the key input.
The output features after each block vary in scale. 

We use self-attention, max-pooling and MLP encoder to encode the trajectory for each agent, followed by another self attention layer with residual skips to keep the interaction between the agents.

We adopt multi-head self-attention model to fuse the features from flow and occupancy branches followed by 8 Cross-Attention modules for 8 future timesteps to better combine each grid with the trajectory data associated with this grid.
Again, we used the occupancy and flow features as queries for the attention module. 
The motion features are used as keys and values.

\textbf{OFMPNet-Swin Decoder.} To refine and extract the output information we use a spatial convolutional decoder which includes multiple 2D convolutions with residual blocks followed by feature pyramid network consisting of 3D spatial convolutions. 
We split the decoder for occupancy and flow branches to better extract the features for each task. 
The output of the occupancy branch for each timestep is a 2D grid including the observed occupancy $\hat{O}_k^b$ and the occluded one $\hat{O}_k^c$. While the output of the flow branch is a 2D grid including the flow $\hat{F}_k$ across $x$ and $y$ axis.

\subsubsection{OFMPNet-R2AttU-T2}
We design a dual recurrent residual convolutional neural network based on U-Net encoder-decoder architecture with attention mechanism named (R2AttU) as a baseline for occupancy and flow prediction task. See Figure \ref{fig:ofmpnet} (OFMPNet-R2AttU-T2).
This architecture is suitable for occupancy prediction task only as it slowly converges for flow prediction task.
\subsubsection{OFMPNet-ULSTM}
Next, we replace the residual convolutional layers in (R2AttU) with LSTM blocks to capture the flow features in our model (ULSTM). See Figure \ref{fig:ofmpnet} (OFMPNet-ULSTM).
During experiments, we add additional separate prediction heads for occupancy and flow tasks (OFMPNet-ULSTM-H) 

\subsection{Occupancy and Flow Loss}
We follow \cite{33da240162c1420494dfbe83c72095d6}, to train the model in a supervised manner. 
We use a per grid binary logistic cross-entropy loss as an occupancy loss between the predicted grids $\hat{O}_k$ and GT grids $O_k$ as follows:
\begin{equation}
    L_O = \sum_{t= 1}^{T_f} \sum_{x=0}^{W-1} \sum_{y=0}^{H-1} CE(\hat{O}_t(x,y), O_t(x,y)),
\end{equation}
where $CE$ refers to the cross-entropy loss.

\begin{figure*}
    \begin{center}
        \includegraphics[width=0.8\linewidth]{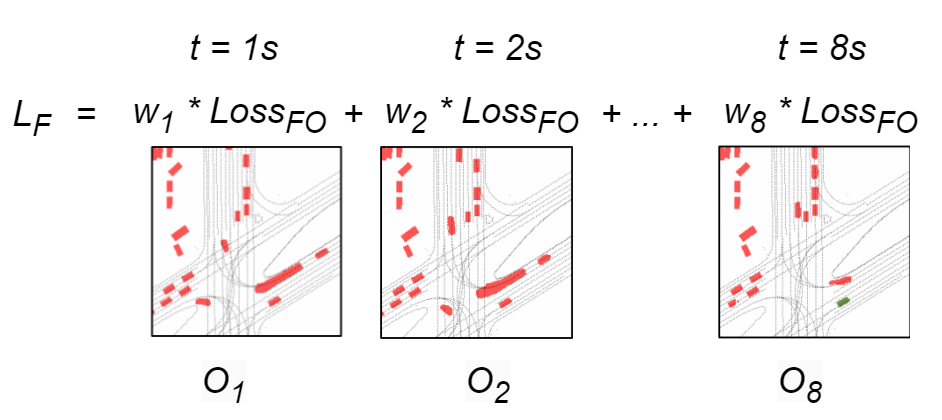}
    \end{center}
    \caption{Time-based weight $w_t$ for flow loss}
    \label{fig:flow_loss}
\end{figure*}

We use a weighted $L1$-norm regression loss as a flow loss between the predicted flow $\hat{F}_k$ and GT flow $F_k$. For better guidance for flow prediction with further timestamps in the future, we add a time-based weight $w_t$ for this flow loss as shown in Figure \ref{fig:flow_loss} . The weight coefficient is $O_k$ as follows:
\begin{equation}
    L_F =  \sum_{t= 1}^{T_f} \sum_{x=0}^{W-1} \sum_{y=0}^{H-1} w_t||\hat{F}_t(x,y) - F_t(x,y) ||_1 O_t(x,y).
\end{equation}

Additionally, we use auxiliary loss (Flow Trace Loss) to benefit from the flow information in occupancy task.
From the input occupancy (current frame), we can use current occupancy $O_0$ with first flow $F_1$ at $t=1$ to construct a grid of future occupancies at $t=1$ and recursively, we apply this process as:
\begin{equation}
    W_t = F_t \circ W_{t-1},
\end{equation}
where $W_t$ is the flow-warped occupancy at $t$ and $W_0 = O_0$.
We multiply each $W_t$ with its occupancy prediction $\hat{O}_t$ to match the GT Occupancy as follows:

\begin{equation}
    L_W = \sum_{t= 1}^{T_f} \sum_{x=0}^{W-1} \sum_{y=0}^{H-1} H(W_t(x,y)\hat{O}_t(x,y), O_t(x,y)).
\end{equation}

Therefore, we define our implemented loss is a combination of these three losses:
\begin{equation}
    L = \sum_K \frac{1}{HWT_f} (\lambda_O L_O + \lambda_F L_F + \lambda_W L_W),
\end{equation}
where $\lambda_O,\lambda_F,\lambda_W$ are coefficients and $K$: all classes.

\subsection{Evaluation metrics}

\subsubsection{Occupancy Metrics}

The occupancy metrics are used for the first and second tasks. 
They compare ground-truth and predicted occupancy grids for each of the $N$ waypoints. 
Let $O_k^b, \hat{O}_k^b$, denote the ground-truth and predicted future occupancy at waypoint $k$ of currently-observed vehicles. 
Let $O_k^c, \hat{O}_k^c$, denote the ground-truth and predicted future occupancy at waypoint $k$ of currently-occluded vehicles.

Treating the occupancy of each grid cell as a separate binary prediction, the $AUC$ metric uses a linearly-spaced set of thresholds in $[0, 1]$ to compute pairs of precision and recall values and estimate the area under the PR-curve. 
More specifically, we compute $AUC(O_k^b, \hat{O}_k^b)$ for currently-observed vehicles and $AUC(O_k^c, \hat{O}_k^c)$ for currently-occluded vehicles.

The \textit{Soft-IoU} metric measures the soft intersection-over-union between ground-truth and predicted occupancy grids as:
\begin{equation}
    \textit{Soft-IoU}(O_k^b,\hat{O}_k^b) = \frac{\sum{O_k^b \hat{O}_k^b}} {\sum{O_k^b + \hat{O}_k^b + O_k^b \hat{O}_k^b}}. 
\end{equation}
If $O_k^b$ is empty, the \textit{Soft-IoU} metric is set to zero. Similarly, we compute \textit{Soft-IoU}$(O_k^c,\hat{O}_k^c)$ for currently-occluded vehicles.

All metrics are averaged over the $N$ predicted waypoints.

\subsubsection{Flow Metrics}

The End-Point Error ($EPE$) metric measures the mean $L2$ distance between the ground-truth flow field $F_k$ and predicted flow field $\hat{F}_k$
 as $||F_{k}(x,y)-\hat{F}_{k}(x,y)||_2$,   where  $F_k(x,y) \neq (0,0)$.

\subsubsection{Joint Occupancy and Flow Metrics}

The joint metrics measure the joint accuracy of occupancy and flow predictions at each waypoint $k$. Given three predictions $\hat{O}_k^b, \hat{O}_k^c, \hat{F}_k$ for waypoint $k$, we compute the Flow-Grounded Occupancy metrics as follows:
First, we compute the ground-truth occupancy of all vehicles (currently observed or occluded) at waypoint $k$ as:
\begin{equation}
    O_k = O_k^b + O_k^c
\end{equation}
and at waypoint $k - 1$ as
\begin{equation}
    O_{k-1} = O_{k-1}^b + O_{k-1}^c.
\end{equation}
We also compute the predicted occupancy of all vehicles as
\begin{equation}
    \hat{O}_k = \hat{O}_k^b + \hat{O}_k^c.
\end{equation}
If the predicted occupancies are accurate, we should have $\hat{O}_k = O_k$. 
The occupancy metrics defined above already evaluate this expectation. 
To ensure correctness of the predicted flow field, $\hat{F}_k$, we use it to warp the ground-truth origin occupancy of that flow field ($O_{k-1}$) as
\begin{equation}
    \hat{W}_k = \hat{F}_k \circ O_{k-1},
\end{equation}
where $o$ indicates function application -- applying the flow field as a function to transform the occupancy. 
If the predicted flow is accurate, it should be able to reach and cover the future occupancy $O_k$. 
Note that since we predict backward flow fields, $\hat{W}_k$ may predict expansion of occupancy in different directions and reach a larger area beyond $O_k$. 
Therefore, we multiply $\hat{W}_k$ element-wise with $\hat{O}_k$, to get $\hat{W}_k\hat{O}_k$.

If the predicted occupancy and flow at waypoint $k$ are accurate, this term should be equal to the ground-truth $O_k$. 
In other words, for a grid cell to be marked as occupied in $\hat{W}_k\hat{O}_k$, it should be supported by both occupancy and flow predictions. 
Therefore, the flow-grounded occupancy metrics compute $AUC$ and \textit{Soft-IoU} between $\hat{W}_k\hat{O}_k$ and ground-truth $O_k$ as $AUC(O_k, \hat{W}_k\hat{O}_k)$ and \textit{Soft-IoU}$(O_k, \hat{W}_k\hat{O}_k)$.
All metrics are averaged over the $N$ predicted waypoints

\section{Experiments}

\subsection{Dataset}

We use Waymo Open Motion dataset (WOD) \cite{ettinger2021large} to train and validate our model. 
WOD contained more than 500k trainng and testing samples collected from real-world road and traffic scenarios. 
WOD include three main labeled classes: vehicle, pedestrian and cyclist.

WOD is split into  485,568 samples as train set,  4,400 samples for validation set and 4,400 samples for test set.
The WOD motion dataset contains agent positions (and other attributes) at 10 Hz. 
Each example in the training and validation sets contains 1 second of history data ($T_h = 1$), 1 timestep for the current time, and 8 seconds of future data ($T_f = 8$). 
This corresponds to 10 history timesteps, 1 current timestep, and 80 future timesteps, for a total of 91 timesteps per scene. 
The 91 timesteps in each example can be organized as:
$[t - 10, t - 9, …, t - 1, t, t + 1, …, t + 79, t + 80],$
where $t$ indicates the current timestep. 
Agents are not necessarily present in all timesteps. 
For example, an agent may get occluded in the future, or it may be currently occluded and only appear in future data.
Presence and absence of agents at particular timesteps is indicated by the valid data attributes.

The test set hides the ground truth future data. 
Each example in the test set contains a total of 11 timesteps (10 history and 1 current timesteps).

In WOD, grid resolution is $256\times256$ convering and area of $80\times80 m^2$.
We consider up to $n = 64$ agents in the scene.

\subsection{Waypoints}
When constructing the ground-truth for the occupancy flow challenge, we divide the 80 future timesteps into 8 one-second intervals, each containing 10 timesteps. 
For example, the first interval consists of the first 10 future timesteps $[t + 1, t + 2, …, t + 10]$ and the last waypoint consists of the last 10 future timesteps $[t + 71, t + 72, …, t + 80]$ in every scenario.

For all tasks in Waymo Occupancy and Flow Prediction challenge the objective is to predict 8 disjoint BEV grids corresponding to the 8 intervals. 
The prediction targets are taken from the last timestep in each interval. 
Ground-truth flow fields are constructed from the displacements between the waypoints.

In ground truth, flow is constructed between timesteps which are 1 second (10 steps) apart. 
For example, flow for the last timestep $t + 80$ is constructed by comparing timestep $t + 80$ with timestep $t + 70$.
Every occupied grid cell at timestep $t + 80$ stores a $(dx, dy)$ vector pointing to its earlier location at timestep $t + 70$. 
The flow vectors point back in time. 

\subsection{Training Setup}

We mainly trained out model on Waymo Open Motion Dataset, where $T_h = 10$ or one past second sampled at $10Hz$ and $T_F = 8$ or eight future seconds sampled at $1Hz$. 
We use Adam optimizer, a dropout rate of $0.1$, and learning rate of $1e-4$ with per 3-epochs decay scheduler with a factor of $0.5$. 
We trained our model for $15$ epochs on $6$ Tesla V100 with a batch size of $3$.

\begin{figure*}
    \centering
     \begin{subfigure}[h]{1\textwidth}
         \centering
         \includegraphics[width=1\linewidth]{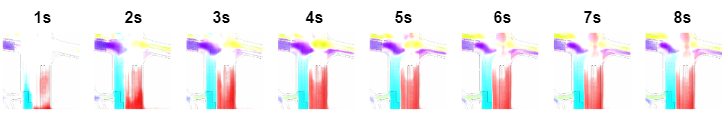}
         \caption{OFMPNet-R2AttU-T2}
         \label{a}
     \end{subfigure}
     \hfill
     \begin{subfigure}[h]{1\textwidth}
         \centering
         \includegraphics[width=1\linewidth]{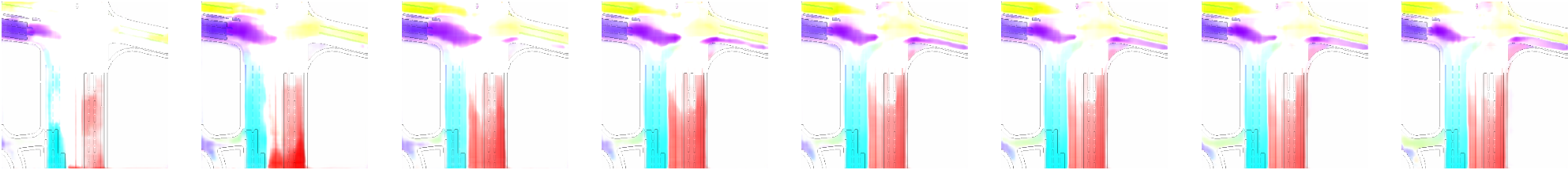}
         \caption{OFMPNet-ULSTM}
         \label{b}
     \end{subfigure}
     \hfill
     \begin{subfigure}[h]{1\textwidth}
         \centering
         \includegraphics[width=1\linewidth]{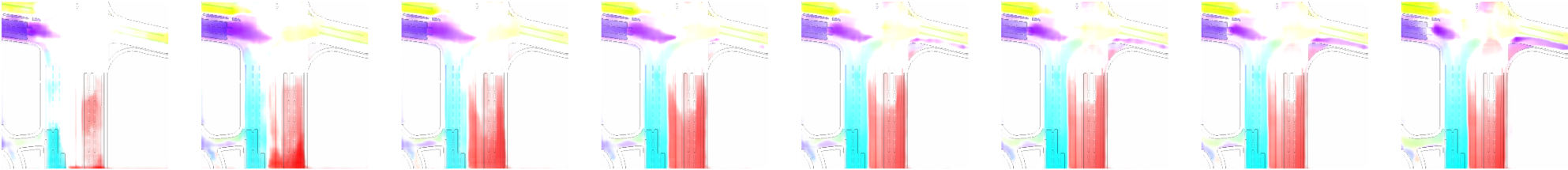}
         \caption{OFMPNet-Swin-T-LSTM}
         \label{c}
     \end{subfigure}
     \hfill
     \begin{subfigure}[h]{1\textwidth}
         \centering
         \includegraphics[width=1\linewidth]{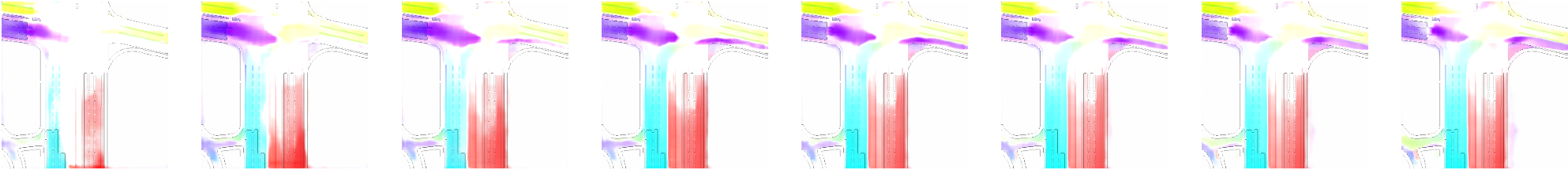}
         \caption{OFMPNet-Swin-T}
         \label{d}
     \end{subfigure}
        \caption{Comparison between OFMPNet flow prediction results on WOM Val-Set}
        \label{fig:compare_visual}
\end{figure*}

\begin{table*}[ht]
    \centering
    \tiny
    \begin{tabular}{lccccccc}
    \hline
    ~ & \multicolumn{2}{c}{Observed Occupancy} & \multicolumn{2}{c}{Occluded Occupancy} & Flow & \multicolumn{2}{c}{Flow-Grounded Occupancy}\\
    \cline{2-8}
    Method             & AUC       & Soft IoU  & AUC       & Soft IoU  & EPE       & AUC       & Soft IoU   \\ 
    \hline
    OFMPNet-R2AttU     &    0.4726 &    0.2028 &    0.0330 &    0.0047 &   21.6873 &    0.5182 &    0.2220  \\
    OFMPNet-ULSTM      &    0.6559 &    0.4007 &    0.1227 &    0.0261 &   20.5876 &    0.5768 &	0.4280  \\
    OFMPNet-ULSTM-H    &    0.6572 &    0.4097 &    0.1180 &    0.0221 &   20.1906 &    0.5835 &	0.4312  \\ 
    OFMPNet-UNext-H    &    0.7119 &    0.4257 &    0.1451 &    0.0309 &   21.6873 &    0.5691 &    0.4243  \\
    OFMPNet-LSTM       &    0.7636 &	0.4910 &	0.1587 &	0.0365 &	3.6859 &	0.7568 &	0.5270  \\
    OFMPNet-CA-LSTM    &    0.7647 &	0.4977 &    0.1583 &	0.0366 &	3.6292 &	0.7594 &	0.5315  \\
    OFMPNet-Swin-T-WL  &    0.7618 &    0.4820 &    0.1540 &    0.0357 & \bf3.3987 & \bf0.7685 &    0.5240  \\
    OFMPNet-Swin-T     & \bf0.7714 & \bf0.5047 & \bf0.1613 & \bf0.0413 &    3.5425 &    0.7621 & \bf0.5410  \\
    \hline
    \end{tabular}
    \caption{Main metrics on Waymo Occupancy and Flow Benchmark (Validation Set)}
    \label{tab:valset}
\end{table*}
\subsection{Ablation Study}

We design our pipeline with different model architectures to ensure choosing the most suitable architecture which provides the best performance in term of metrics and inference time.

With LSTM blocks we achieve +20\% IOU, +15\% AUC for occupancy task and +7\% AUC, +20\% IOU flow task compared to UNet architecture. 
Adding a separate head (H) for each task boosts the model results by 1-3\%.
For further improvement, we use UNext architecture \cite{valanarasu2022unext} which provides +8\% boost for occupancy task. To tackle the difficulty of flow task converge, we use Swin Transformer as explained in details in method section.
Table \ref{tab:valset}, shows the main metrics on Waymo Open Motion Dataset (validation set) for each model in details.
Our model Swin-T-WL which is trained with time-weighted loss show the best AUC flow-grounded occupancy. CA-LSTM represents model with LSTM-based Cross Attention module. UNext-H represents UNext model with additional separate heads for both occupancy and flow tasks.
Table \ref{tab:perf} shows the inference performance of our implemented models on NVidia RTX-3060 GPU in seconds.

Figure \ref{fig:compare_visual} presents crossroad scenario, where we visualizes comparison between ULSTM results (upper figure) and Swin-T results (lower figure) on Waymo Open Motion Validation Set. Each subplot displays the result of 1) observed occupancy, 2) backward flow. Our ULSTM-model works better for short timestamps in the future (2-4seconds), while Swin-T shows more accurate flow prediction for further timestamps in the future (6-8seconds).  

\begin{table*}[ht]
    \scriptsize
    \centering
    \begin{tabular}{lc}
    \hline
    Method             &  Average Latency(s)    \\ 
    \hline
    OFMPNet-R2AttU     &    0.172 $\pm$ 0.023   \\
    OFMPNet-ULSTM      &    0.276 $\pm$ 0.026   \\ 
    OFMPNet-UNext      &    0.211 $\pm$ 0.031   \\
    OFMPNet-Swin-T     &    0.187 $\pm$ 0.019   \\
    \hline
    \end{tabular}
    \caption{Main inference latency of our models on NVidia RTX-3060 GPU}
    \label{tab:perf}
\end{table*}

\begin{table*}[ht]
    \centering
    \tiny
    \begin{tabular}{lccccccc}
    \hline
    ~ & \multicolumn{2}{c}{Observed Occupancy} & \multicolumn{2}{c}{Occluded Occupancy} & Flow & \multicolumn{2}{c}{Flow-Grounded Occupancy}\\
    \cline{2-8}
    Method           & AUC       & Soft IoU  & AUC       & Soft IoU  & EPE       & AUC       & Soft IoU  \\ 
    \hline
    OFMPNet-ULSTM     &    0.6485 &	  0.3823 &	  0.1242 &    0.0230 &   20.0771 &    0.5799 &    0.4070 \\ 
    OFMPNet-R2AttU-T2 &    0.4759 &	  0.2006 &	  0.0403 &    0.0065 &   21.5577 &    0.4846 &    0.2008 \\
    OFMPNet-CA-LSTM   &    0.7627 &   0.4950 &    0.1633 &    0.0374 &    3.6686 &	  0.7590 &    0.5284 \\
    OFMPNet-Swin-T-WL &    0.7591 &   0.4786 &	  0.1618 &    0.0371 & \bf3.4109 & \bf0.7675 &	  0.5207 \\
    OFMPNet-Swin-T    & \bf0.7694 &\bf0.5021 & \bf0.1651 & \bf0.0423 &    3.5868 &    0.7614 & \bf0.5377 \\
    \hline
    \end{tabular}
    \caption{Main metrics on Waymo Occupancy and Flow Benchmark (Testset) }
    \label{tab:testset}
\end{table*}

\begin{figure*}
    \begin{center}
        \includegraphics[width=1\linewidth]{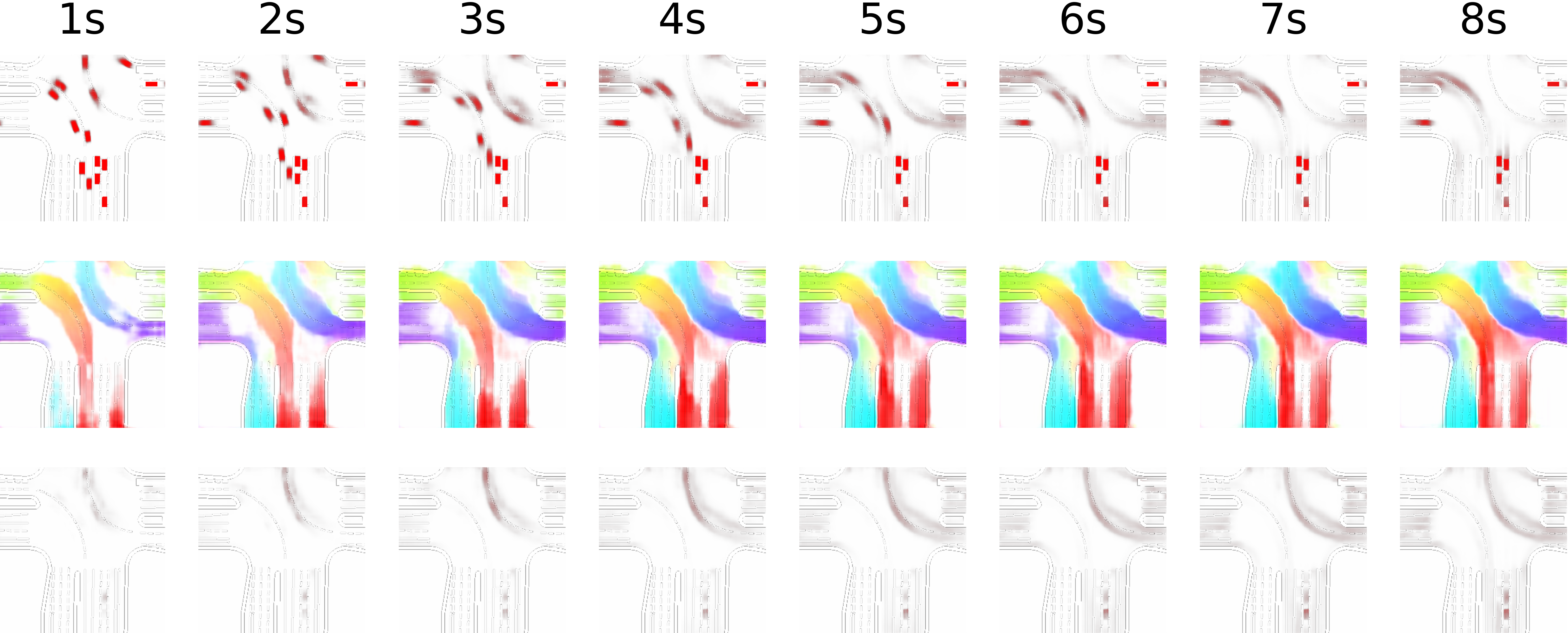}
        \includegraphics[width=1\linewidth]{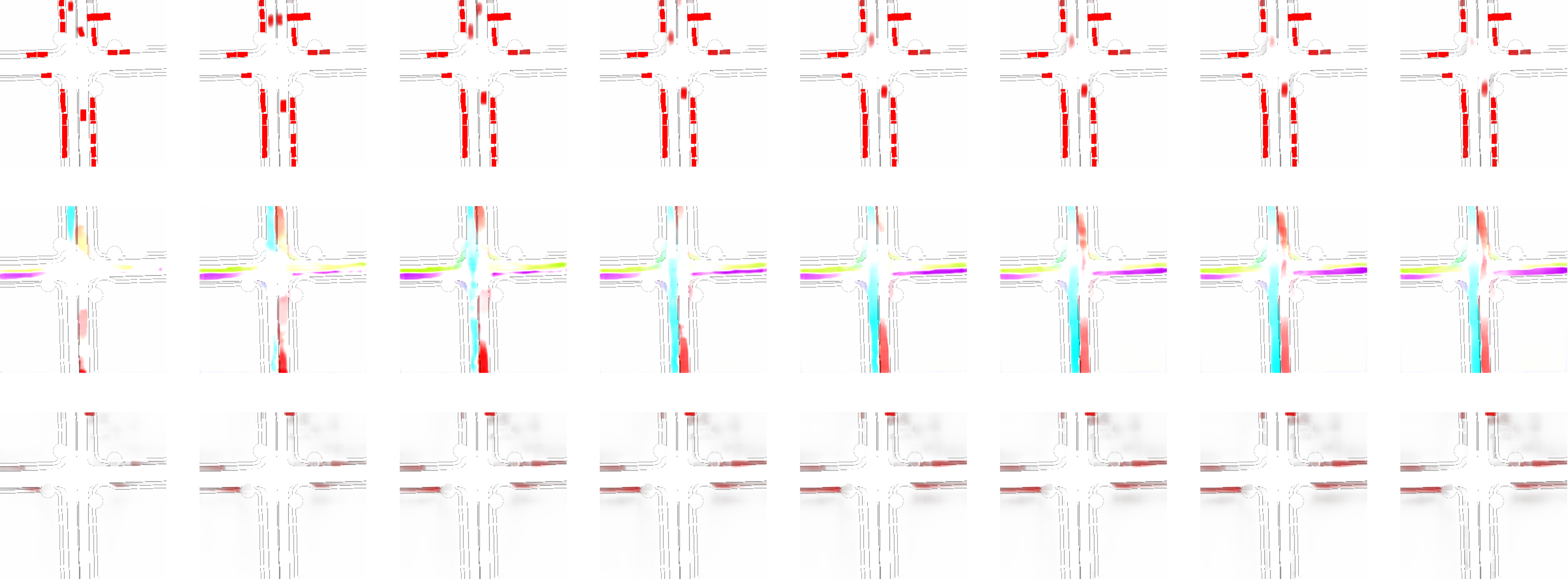}
        \includegraphics[width=1\linewidth]{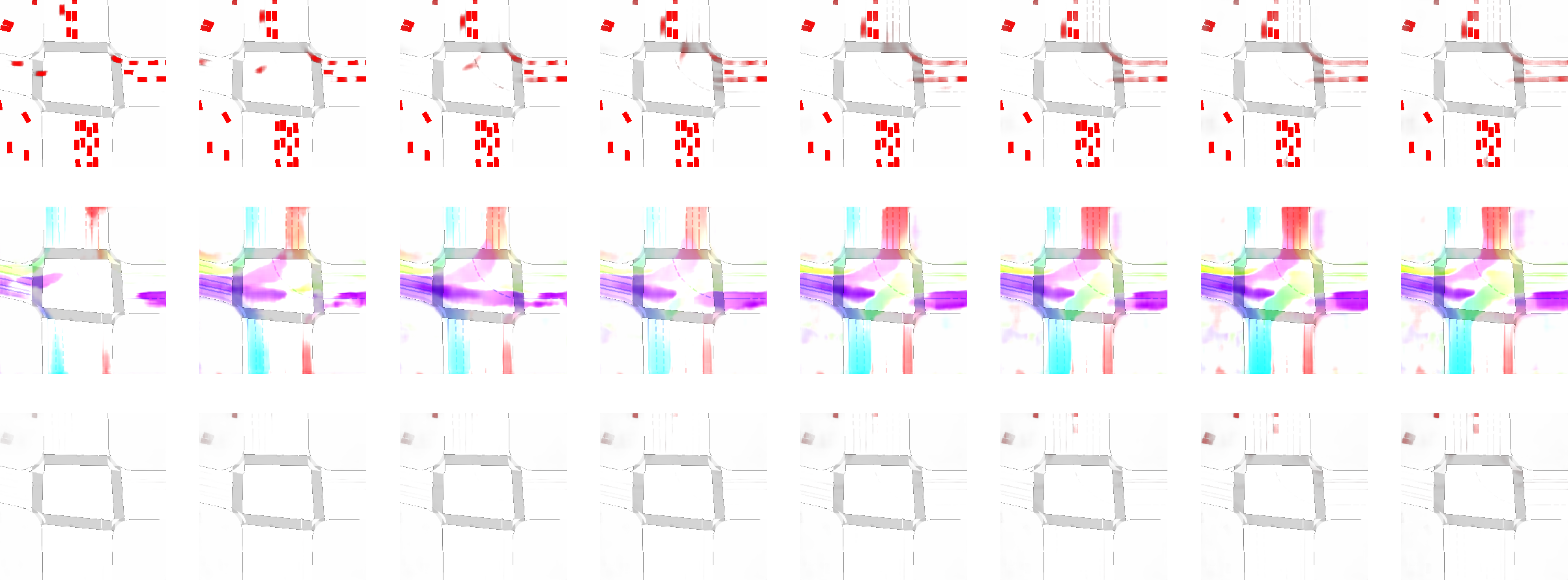}
    \end{center}
    \caption{Qualitative results from Waymo Open Motion Dataset using our pretrained OFMPNet-Swin-T model}
    \label{fig:visual}
\end{figure*}

\subsection{Results on Waymo Open Motion Dataset}
Table \ref{tab:testset} shows the main metrics on Waymo Open Motion Dataset (Test set) for each model. 
Our best model with Swin-Transformer achieves ~77\% AUC on observed occupancy task. Swin-Transformer with weighed flow loss (WL) achieves 76.75\% on flow task (main metric). 

Table \ref{tab:benchmark} shows the public leaderboard for Waymo challenge. In comparison with the-state-of-the-art methods, our method shows the best SOFT IoU metric with 3\% margin and the second best AUC metric for observed occupancy. Our best configuration achieved 76.75\% AUC on Flow-grounded occupancy and demonstrated state-of-the-art results with 50.2\% Soft-IoU.

The trained model, on the same data at different launches, gives the same result, the tables show averaged quality metrics on a large-scale dataset (WOMD), which reflects their significance, such validation is generally recognized in the literature.

Figure \ref{fig:visual} shows qualitative results from Waymo Open Motion Dataset using our pretrained OFMPNet-Swin-T model to intuitively evaluate the performance of our method. Each subplot displays the result of 1) observed occupancy, 2) occluded occupancy, and 3) flow-traced occupancy.
We present different road scenarios including T-intersection, crossroad and ring road. The results show that our method can perform effective and accurate occupancy
forecasting for both dynamic and static traffic agents.

\begin{table*}[ht]
    \centering
    \tiny
    \begin{tabular}{lccccccc}
    \hline
    ~ & \multicolumn{2}{c}{Observed Occupancy} & \multicolumn{2}{c}{Occluded Occupancy} & Flow & \multicolumn{2}{c}{Flow-Grounded Occupancy}\\
    \cline{2-8}
    Method                                  &    AUC    & Soft IoU  &    AUC    & Soft IoU  &    EPE    &   AUC     & Soft IoU  \\ 
    \hline
    \hline
    FTLS \cite{waymo_leaderboard}           &   0.6179  &    0.3176 &   0.0850  &    0.0191 &    9.6118 &   0.6885  &    0.4305 \\ 
    Motionnet \cite{wu2020motionnet}        &   0.6941  &    0.4114 &   0.1413  &    0.0309 &    4.2751 &   0.7324  &    0.4691 \\ 
    3D-STCNN \cite{8788794}                 &   0.6909  &    0.4123 &   0.1153  &    0.0211 &    4.1810 &   0.7333  &    0.4684 \\ 
    Spatial Temp-Conv\cite{waymo_leaderboard}&   0.7438  &    0.2173 &   0.1680  &    0.0189 &    3.8705 &   0.7653  &    0.5376 \\
    VectorFlow\cite{huang2022vectorflow}    &   0.7548  &    0.4884 &\bf0.1736  & \bf0.0448 &    3.5827 &   0.7669  &    0.5298 \\
    STrajNet \cite{liu2022strajnet}         &   0.7514  &    0.4818 &   0.1610  &    0.0183 &    3.5867 &   0.7772  &    0.5551 \\
    Temporal Query \cite{waymo_leaderboard} &   0.7565  &    0.3934 &   0.1707  &    0.0404 &    3.3075 &   0.7784  &    0.4654 \\
    Look Around \cite{waymo_leaderboard}    &\bf0.8014  &    0.2336 &   0.1386  &    0.0285 &    2.6191 &   0.8246  &    0.5488 \\ 
    HorizonOccFlowPredict \cite{waymo_leaderboard}    &\bf0.8033  &    0.2349 &   0.1650  &    0.0169 & \bf3.6717 &\bf0.8389  &  \bf0.6328 \\ 
    \hline
    \bf{OFMPNet (Ours)}                     &   0.7694  & \bf0.5021 &   0.1651  &    0.0423 &    3.5868 &   0.7614  &    0.5377 \\
    \hline
    \end{tabular}
    \caption{Waymo Occupancy and Flow leaderboard}
    \label{tab:benchmark}
\end{table*}

\section{Conclusion}

This paper introduces an end-to-end neural network methodology for predicting the future behaviors of dynamic objects in road scenarios, leveraging the occupancy map and scene flow. We have evaluated our approach, called OFMPNet, which uses a sequence of bird’s-eye-view images, inclusive of a road map, occupancy grid, and prior motion flow. The evaluation involved various model architectures and feature fusion techniques, including Swin-T, Attention, and LSTM. Additionally, we proposed a novel time-weighted motion flow loss to minimize end-point error.

An important advantage of the proposed approach is that its performance does not depend on the number of objects whose movement needs to be predicted. This is ensured by generating dense maps of object motion (occupancy flow) at the model output.

On the Waymo Occupancy and Flow Prediction benchmark, our approach has achieved state-of-the-art results, with a Soft IoU of 50.2\% and an AUC of 76.9\% on Flow-Grounded Occupancy.

The primary limitation of this methodology is its reliance on HD maps during both training and inference phases. Consequently, future work could involve designing a similar approach that is independent of HD maps, thereby making the approach more general and suitable for production. Accelerating this methodology to function on on-board devices with online performance could further enhance its production-friendliness.

\section{Acknowledgment}
This work was partially supported by the Analytical Center for the Government of the Russian Federation in accordance with the subsidy agreement (agreement identifier 000000D730321P5Q0002; grant  No. 70-2021-00138).

\bibliography{ofmpnet}

\end{document}